\documentclass{article}

\usepackage[nonatbib, final]{nips_2018}

\usepackage[utf8]{inputenc} 
\usepackage[T1]{fontenc}    
\usepackage{hyperref}       
\usepackage{url}            
\usepackage{booktabs}       
\usepackage{amsfonts}       
\usepackage{nicefrac}       
\usepackage{microtype}      
\usepackage{mathtools}
\usepackage{multirow}

\usepackage[backend=biber,style=apa, backref=true, sortcites=true,sorting=nyt, uniquename=false]{biblatex}
\DeclareLanguageMapping{american}{american-apa}
\usepackage[american]{babel}
\title{Understanding Multi-Step Deep Reinforcement Learning: A Systematic Study of the DQN Target}

\author{
  J. Fernando Hernandez-Garcia \\
  Department of  Computing Science \\
  University of Alberta \\
  Edmonton, AB, T6G 2E8 \\
  {jfhernan@ualberta.ca} 
  \And
  Richard S. Sutton \\
  Department of Computing Science \\
  University of Alberta \\
  Edmonton, AB, T6G 2E8 \\
  {rsutton@ualberta.ca}
}

\begin{document}

\maketitle

\begin{abstract}
Multi-step methods such as Retrace($\lambda$) and $n$-step $Q$-learning have become a crucial component of modern deep reinforcement learning agents.
These methods are often evaluated as a part of bigger architectures and their evaluations rarely include enough samples to draw statistically significant conclusions about their performance.
This type of methodology makes it difficult to understand how particular algorithmic details of multi-step methods influence learning. 
In this paper we combine the $n$-step action-value algorithms Retrace, $Q$-learning, Tree Backup, Sarsa, and $Q(\sigma)$ with an architecture analogous to DQN.
We test the performance of all these algorithms in the mountain car environment; this choice of environment allows for faster training times and larger sample sizes.
We present statistical analyses on the effects of the off-policy correction, the backup length parameter $n$, and the update frequency of the target network on the performance of these algorithms.
Our results show that (1) using off-policy correction can have an adverse effect on the performance of Sarsa and $Q(\sigma)$; (2) increasing the backup length $n$ consistently improved performance across all the different algorithms; and (3) the performance of Sarsa and $Q$-learning was more robust to the effect of the target network update frequency than the performance of Tree Backup, $Q(\sigma)$, and Retrace in this particular task.
\end{abstract}

A crucial choice when designing reinforcement learning agents is the definition of the update target.
The choice of target heavily influences the bias and the variance of the algorithm, as well as the convergence guarantees and convergence speed of the estimates of the value function. 
This choice is particularly overwhelming given the large number of methods that have been proposed: should one use Sarsa or $Q$-learning; should one use off-policy correction; if so, should one truncate the importance sampling ratio.
Systematic studies of all these algorithms are essential to be able to make informed decisions when designing reinforcement learning agents.

A particular innovation with long lasting influence in reinforcement learning research is the TD($\lambda$) algorithm \parencite{sutton1988learning}.
Through the introduction of the trace decay parameter $\lambda$, TD($\lambda)$ unifies Monte Carlo estimation and one-step temporal difference methods.
These type of methods are now known as \textit{multi-step} methods and they represent a large family of algorithms that one could choose from when designing reinforcement learning agents.

Multi-step methods are often encountered in one of two formulations: $n$-step methods which use trajectories of length $n$ to compute the target, and eligibility trace methods which, like the TD($\lambda$) algorithm, combine the information of many $n$-step targets to create a compound target.
These types of methods have been exhaustively studied, both theoretically and empirically, in the tabular and linear function approximation case.
In deep reinforcement learning research, where the value function is approximated using a neural network, multi-step methods have seen a resurgence in recent years.
Algorithms such as Retrace($\lambda$) \parencite{remi_retrace} are now a key component of large-scale distributed architectures such as Rainbow \parencite{rainbow2017}, Reactor \parencite{reactor2017} and IMPALA \parencite{impala2018}.
Other architectures, such as Ape-X \parencite{apex2018}, use an $n$-step variant of $Q$-Learning \parencite{watkins1989qlearn} to estimate the action-value function.
Impressively, even though the $n$-step $Q$-Learning target does not use off-policy corrections, the architecture is still capable of remarkable results.

Despite being widely used in deep reinforcement learning, direct comparisons of multi-step methods are scarce.
Even when comparisons are available, researchers often omit reporting standard deviations, which makes it difficult to draw statistically significant conclusions about the performance of these algorithms.
Moreover, several mutli-step methods still remain unexplored in the deep reinforcement learning literature.

The main reason significance has taken such a secondary role in deep reinforcement learning research is that evaluations are often done in high-dimensional environments where training is computationally expensive.
In this paper we scale down to the relatively small environment mountain car to obtain significant comparisons of all these algorithms under different parameter settings.
Moreover, we test two algorithms that have not been previously tested in the deep reinforcement learning setting: Sarsa and $Q(\sigma)$.
We implemented the $n$-step versions of the algorithms Retrace($\lambda$), $Q$-Learning, Tree Backup \parencite{precup2000}, Sarsa \parencite{rummery1995}, and $Q(\sigma)$ \parencite{qsigma-paper} each combined with an architecture analogous to DQN. 
We then tested the effect of the off-policy correction, the parameter $n$, and the update frequency of the target network on the performance of each of these algorithms.

\section{Notation}

In reinforcement learning, the sequential decision-making problem is modeled using the Markov Decision Process formulation defined by the tuple $\langle \mathcal{S}, \mathcal{A}, \mathcal{R}, P, \gamma \rangle$.
In this formulation, an agent --- which is both a learner and an actor --- and an envi`ronment interact over a sequence of discrete time steps $t \geq 0$. 
At every time step the agent receives a state $S_t \in \mathcal{S}$, which encodes information about the environment.
Based on that information, the agent chooses and executes an action $A_t \in \mathcal{A}$. 
As a consequence of the executed action, the environment sends back to the agent a new state $S_{t+1}$ and a reward $R_{t+1} \in \mathcal{R}$.
The reward and new state are distributed according to the transition dynamics probability function $P$, which defines the probability of observing $(S_{t+1}, R_{t+1})$ given $(S_t, A_t)$.

Actions are selected according to a policy $\pi(\cdot | S_{t})$ --- a probability distribution over the actions given the current state.
The goal of the agent is to maximize the expected sum of discounted rewards $\mathbb{E}_\pi \big\{ \sum_{i \geq t} \gamma^{i - t} R_{i+1} \big\}$, where the expectation is with respect to $\pi$ and $P$, and $\gamma$ is a discount factor in the interval $[0,1]$. 
In order to make informed decisions, agents often estimate the action-value function $q_\pi$ which maps states and actions to a value in $\mathbb{R}$ and is defined as
\begin{equation}
\label{eq:qpi}
    q_\pi (s,a) \overset{.}{=} \mathbb{E}_\pi \Big \{ \sum_{i \geq t} \gamma^{i-t} R_{i+1} \big | S_t = s, A_t = a \Big \}.
\end{equation}
To approximate the action-value function, agents use update rules based on stochastic approximation algorithms of the form:
\begin{equation}
    Q_{t+1}(S_t, A_t) \leftarrow Q_{t}(S_t, A_t) + \alpha [\hat{G}_t - Q_{t}(S_t, A_t) ], t \geq 0,
\end{equation}
where $\hat{G}_t$ is an estimate of the sum of discounted rewards, also known as the \textit{target} of the update function, and $Q_t$ is the current estimate of the action-value function. 

\section{n-Step Methods}

Algorithms for estimating the action-value function are often characterized by the target of their update function.
For example, the one-step Sarsa algorithm uses the target:
\begin{equation*}
    \hat{G}_t \overset{.}{=} R_{t+1} + \gamma Q_t(S_{t+1}, A_{t+1}).
\end{equation*}
This target estimates the sum of discounted reward by bootstrapping on the estimate of the future reward $Q_t(S_{t+1}, A_{t+1})$.
If $Q_t(S_{t+1}, A_{t+1})$ was exactly $q_\pi(S_{t+1}, A_{t+1})$, then the expected value of $\hat{G}$ with respect to $\pi$ and $P$ would be exactly $q_\pi$.
If $Q_t$ is not exactly $q_\pi$, then the target is slightly biased \parencite{Kearns:2000:BEB:648299.755183}.

One way to reduce this type of bias is to take longer trajectories that contain more observations of future reward.
If we apply this to the Sarsa algorithm, we obtain the $n$-step target:
\begin{equation}
    \hat{G}_{t:t+n} \overset{.}{=} R_{t+1} + \gamma R_{t+2} + ... + \gamma^n Q_t(S_{t+n}, A_{t+n}),
\end{equation}
where $n$ is also known as the backup length parameter.
We can simplify this expression by defining the estimate of the return recursively:
\begin{align}
    \label{eq:nstep_sarsa}
    \hat{G}^S_{t:t+n} = R_{t+1} + \gamma \hat{G}^S_{t+1:t+n}, \hspace{50pt}
    \hat{G}^S_{t+n:t+n} = Q_t(S_{t+n}, A_{t+n})
\end{align}

Similar to Sarsa, other $n$-step algorithms can also be defined recursively.
For example, the target of the $n$-step Tree Backup algorithm \parencite{precup2000} can be defined as
\begin{equation}
    \label{eq:nstep_treebackup}
    \hat{G}^{TB}_t = R_{t+1} + \gamma \Big[ \pi(A_{t+1}|S_{t+1}) \hat{G}^{TB}_{t+1:t+n} 
                    + \sum_{a \neq A_{t+1}} \pi(a | S_{t+1}) Q_t(S_{t+1}, a) \Big],
\end{equation}
where the base case is the same as in $n$-step Sarsa.
The main difference between Sarsa and Tree Backup is that Sarsa samples a new action at every step of the backup whereas Tree Backup takes an expectation over all the possible actions.

$Q(\sigma)$ is a recent algorithm that combines $n$-step Sarsa and Tree Backup \parencite{qsigma-paper}.
In this case the parameter $\sigma$ controls the degree of expectation and sampling at every step of the backup.
The recursive definition of the target of $n$-step $Q(\sigma)$ is
\begin{align}
    \label{eq:nstep_qsigma}
    \hat{G}^\sigma_{t:t+n} &= R_{t+1} + \gamma \Big[ \big(\sigma_{t+1} + (1-\sigma_{t+1})\pi(A_{t+1}|S_{t+1}) \big) 
                            \hat{G}^\sigma_{t+1:t+n} \nonumber \\
    & \hspace{60pt} + 
        (1 - \sigma_{t+1}) \sum_{a \neq A_{t+1}} \pi(a|S_{t+1}) Q_t(S_{t+1}, a) \Big],
\end{align}    
where $\sigma_t \in [0,1]$ for all $t$ and the base case is the same as in $n$-step Sarsa and Tree Backup.
The parameter $\sigma$ makes it possible to reap the benefits of Sarsa and Tree Backup at different steps of training.
A simple heuristic that has been proposed to achieve such benefit is decaying the parameter $\sigma$ from $1$ (Sarsa) to $0$ (Tree Backup) over the course of training. 

\subsection{Off-Policy Algorithms}

All the algorithms introduced so far estimate the correct action values as long as the samples are collected according to the target policy $\pi$; this is known as the \textit{on-policy} case.
Alternatively, samples could be collected according to a different policy $\mu$, known as the behaviour policy, while estimating the action values of policy $\pi$.
This is the case of \textit{off-policy} learning. 

In off-policy learning, the estimates of the return have to be corrected to account for the mismatch in the policies. 
This can be easily done in the $n$-step Sarsa algorithm by using importance sampling. 
In such case, the estimate of the return is defined as
\begin{equation}
    \label{eq:offpolicy_sarsa}
    \hat{G}^S_{t:t+n} = R_{t+1} + \gamma \rho_{t+1} \hat{G}^S_{t+1:t+n}, \hspace{20pt}
    \rho_{t+1} \overset{.}{=} \frac{\pi(A_{t+1}|S_{t+1})}{\mu(A_{t+1}|S_{t+1})}.
\end{equation}
This algorithm is equivalent to the per-decision importance sampling from \citeauthor{precup2000} (\citeyear{precup2000}).

The Tree Backup algorithm does not need to be adjusted since it is already computing the estimates of the action-value function under the correct policy.
On the other hand, $Q(\sigma)$ can be adapted to the off-policy setting by combining per-decision importance sampling and Tree Backup:
\begin{align}
    \label{eq:offpolicy_qsigma}
    \hat{G}^\sigma_{t:t+n} &= R_{t+1} + \gamma \Big[ \big(\sigma_{t+1} \rho_{t+1} 
                            + (1-\sigma_{t+1})\pi(A_{t+1}|S_{t+1}) \big) 
                            \hat{G}^\sigma_{t+1:t+n} \nonumber \\
    & \hspace{60pt} + 
        (1 - \sigma_{t+1}) \sum_{a \neq A_{t+1}} \pi(a|S_{t+1}) Q_t(S_{t+1}, a) \Big].
\end{align}    

In the off-policy case, we will study two more algorithms: $n$-step Retrace and $n$-step $Q$-learning.
Retrace was originally proposed as an eligibility trace algorithm, but it can easily be adapted to the $n$-step case.
In this case, the corresponding target is defined as
\begin{equation}
    \label{eq:nstep_retrace}
    G^R_{t:t+n} = R_{t+1} + \gamma \Big[ c_{t+1} \hat{G}^R_{t+1:t+n} 
                + \sum_{a \in \mathcal{A}} \pi(a|S_{t+1}) Q_t(S_{t+1}, a) - c_{t+1}Q_t(S_{t+1}, A_{t+1})  \Big],
\end{equation}
where $c_{t+1} = \min(k, \rho_{t+1})$, $k$ is a positive cutoff parameter often set to $1$, and the base case is the same as for the previous algorithms.

$n$-step $Q$-learning does not fall exactly within the off-policy family of algorithms since it does not correct for the mismatch between the target and behaviour policy.
Nevertheless, in practice it has shown promising results \parencite{apex2018}; hence, we will also study its performance.
The target of the $n$-step $Q$-learning algorithm is defined as:
\begin{align}
    \label{eq:nstep_qlearning}
    \hat{G}^{QL}_{t:t+n} = R_{t+1} + \gamma \hat{G}^{QL}_{t+1:t+n}, \hspace{50pt} \hat{G}^{QL}_{t+n:t+n} = \max_a Q_t(S_{t+n}, a).
\end{align}
Note that the base case is different than the base case for all the previous algorithms.
In the one-step case, $n$-step $Q$-learning estimates the action-values corresponding to the optimal policy $\pi^*$ corresponding to the optimal action-value function $q_*(s,a) \overset{.}{=} \max_\pi q_\pi(s,a) \ \forall \ (s,a) \in \mathcal{S} \times \mathcal{A}$.
For $n$ greater than $1$, it is not clear what is the policy corresponding to the action-values that are being estimated.

\section{The DQN Architecture}

The action-value function $q_\pi$ can be estimated exactly by storing an individual estimate for each state-action pair. 
When this is not feasible, it can be approximated using a parameterized function $q(\cdot, \cdot, \boldsymbol{\theta}_t)$, where $\boldsymbol{\theta}_t \in \mathbb{R}^d$ is a parameter vector of size $d$.
The weight vector $\boldsymbol{\theta}_t$ can then be learned by using semi-gradient descent \parencite{sutton2018}. 

For high dimensional state spaces, one of the most popular methods for approximating the action-value function is the DQN architecture \parencite{mnih2015humanlevel}. 
In essence, DQN is a neural network that takes in information about the state of the environment and outputs the estimate of the action-value function for each possible action. 
The neural network is trained by minimizing the loss function
\begin{equation}
    \label{eq:dqn_loss}
    l(\boldsymbol{\theta}_t) = \big(R_{t+1} + \max_a q(S_{t+1}, a, \boldsymbol{\theta}^-_t) - q(S_t, A_t, \boldsymbol{\theta}_t) \big)^2,
\end{equation}
with respect to $\boldsymbol{\theta}_t$.
Note that the architecture requires two sets of parameters: $\boldsymbol{\theta}_t$, the set of parameters that are being learned at every time step; and the target network's parameters $\boldsymbol{\theta}^-_t$, which are updated less frequently and used exclusively to compute the target of the loss function.
The update frequency of $\boldsymbol{\theta}^-_t$ is a hyper-parameter known as the target network update frequency.

DQN agents do not learn directly from samples collected from the environment.
Instead, observations are stored in a buffer, known as the experience replay buffer.
At every training step, a mini-batch of observations is sampled uniformly from the buffer to compute an update for $\boldsymbol{\theta}_t$. 
Since the policy at the time of storing might not be equal to the policy at the time of sampling from the buffer, the target of the DQN loss has to be capable of estimating the return off-policy.
This is not a problem in the original DQN since it uses the $Q$-learning target.
However, this fact should be accounted for when combining other methods with this type of architecture. 

Overall, the experience replay buffer introduces three extra hyper-parameters.
First, the replay memory, which is the maximum number of observations that can be stored in the buffer.
Second, the replay start size, which is a number of random actions executed before training starts to populate the buffer. 
Lastly, the mini-batch size, which is the number of observations sampled from the buffer at every training step. 

\subsection{Adapting the DQN Architecture}

For our experiments we adapted the DQN architecture to work with all the algorithms described in the previous section. 
The loss function can be readily modified by substituting the target in the loss function.
For each method, the loss function used by the architecture is
\begin{equation}
    \label{eq:modified_dqn_loss}
    l(\boldsymbol{\theta}_t) = \big(\hat{G}_{t:t+n}(\boldsymbol{\theta}^-_t) - q(S_t, A_t, \boldsymbol{\theta}_t) \big)^2,
\end{equation}
where $\hat{G}_{t:t+n}(\boldsymbol{\theta}^-_t)$ is the corresponding estimate of the return computed using the target network.

The experience replay buffer also needs to be modified.
In the case of the algorithms that compute the importance sampling ratio, the probability of taking an action $\pi_k(A_k|S_k)$ is also stored in the buffer.
Then, at the time of computing the update, the importance sampling ratio is computed as $\frac{\pi_t(A_k|S_k)}{\pi_k(A_k|S_k)}$, where $t$ is the time at which the update is computed and $k$ is the time at which the observation was stored.
In the case of $Q(\sigma)$, the parameter $\sigma_t$ is also stored in the buffer.
Overall, each entry in the buffer is of the form $(S_{k}, A_{k}, R_{k}, \mathbb{I}_k, \pi_k(A_k | S_k), \sigma_k)$, where $R_k$, $\pi_k$, and $\sigma_k$ are set to zero if $k$ corresponds to the first time step of an episode and $\mathbb{I}_k$ is a boolean indicating if $S_k$ is a terminal state.

\section{Empirical Evaluations}

In our empirical evaluations, we study the effect of the off-policy correction, the parameter $n$, and the target network update frequency on the performance of the algorithms introduced in the previous sections.
The results of this systematic study facilitate an in-depth understanding of these algorithmic details which could inform the design of new deep reinforcement learning agents.

In order to test the performance of all these algorithms and to allow for enough samples to obtain statistical significance, we used the mountain car environment as described in \citeauthor{sutton2018} (\citeyear{sutton2018}).
To prevent algorithms from running for too long, we enforced a timeout of 5,000 time steps after which an episode was stopped.
Note that timing out is not treated the same as termination.
In the case of termination, all the subsequent rewards and action-values are considered zero when computing the target.
On the other hand, in the case of a time out, the $n$-step target is truncated at the last available time step effectively making the $n$-step return shorter.
We found in preliminary experiments that treating time outs the same way as terminations can have catastrophic effects in learning.
All the agents were trained for $500$ episodes.

The network architecture remained constant for all the experiments.
It consisted of an input layer, a fully-connected hidden layer, and a fully-connected output layer.
The input layer consisted of a two-dimensional vector with the position and velocity of the car; the fully-connected hidden layer consisted of $1,000$ rectified linear units; and the output layer was a fully-connected linear layer that returned the action-value for each action.
To minimize the loss, we used the RMSprop optimizer \parencite{Tieleman2012} with gradient momentum and squared gradient momentum of 0.95, and a minimum squared gradient of 0.01 --- the same values used in the original DQN. 
For the learning rate parameter $\alpha$ we used a value of $0.00025$ as in the original DQN architecture.

The replay memory size of the architecture was $20,000$ with a replay start size of $1,000$ random actions and a mini-batch size of $32$. 
The default value for the target network update frequency was $1,000$; however, for one of the experiments we investigated the effect of this hyper-parameter on the performance of each algorithm.
We will make special emphasis when using a different value for the target network update frequency.

All the agents behaved and estimated the action-values corresponding to an $\epsilon$-greedy policy with $\epsilon = 0.1$.
Note that even when the behaviour and target policy are the same, the agent needs to compute the importance sampling ratio if it is to correct for the discrepancy between the policy at the time of storing and the policy at the time of sampling.
We used a discount factor $\gamma$ of 1 to encourage agents to find the terminal state before episodes timed out.
We did not anneal $\epsilon$ for any of our experiments since we found in preliminary results that it did not result in a significant difference in performance.

\subsection{Off-Policy vs On-Policy}

Our first experiment was motivated by the results obtained with $n$-step $Q$-learning without off-policy corrections in the Ape-X architecture \parencite{apex2018}.
In light of those results, we investigated how other algorithms would perform without any off-policy correction.
We hypothesized that similar to $n$-step $Q$-learning, Sarsa and $Q(\sigma)$ would not see an adverse effect in their performance if they did not use off-policy correction.

We tested two versions of the one-step algorithms Sarsa, $Q(\sigma = 0.5)$, and decaying $\sigma$: a version with off-policy correction --- called off-policy, and without off-policy correction --- called on-policy.
The decaying $\sigma$ agent is a $Q(\sigma)$ agent that has an initial value of $\sigma$ of 1 which decreases by 0.002 at the end of each episode.
In the case of the off-policy agents, the importance sampling ratio is defined as $\frac{\pi_t(A_k|S_k)}{\pi_k(A_k|S_k)}$, where $\pi_t$ is the policy at the time of the update and $\pi_k$ is the policy at the time of storing the observation in the buffer.

We used the average return per episode as measure of performance.
We studied this performance measure at three different time scales: the first 50 episodes of training (initial performance), the last 50 episodes of training (final performance), and the whole training period (overall performance).
We ran $100$ independent runs for each agent.

\begin{figure}[t]
  \centering
  \label{fig:off_vs_on}
  \includegraphics[width=1\textwidth]{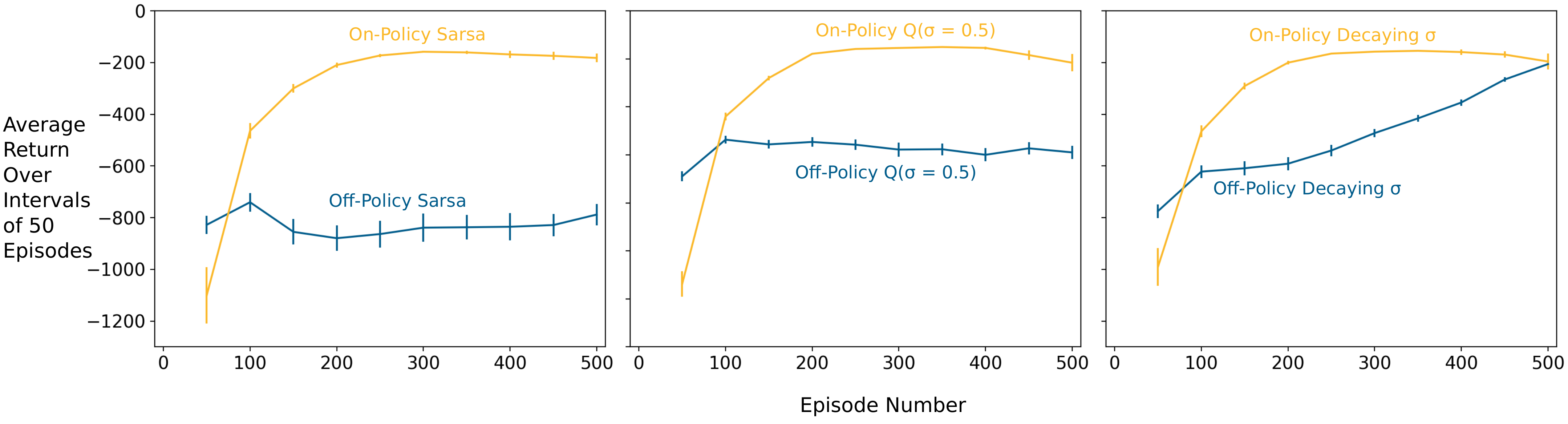}
  \caption{Average return over intervals of 50 episodes. The results are averaged over 100 independent runs. The error bars correspond to 95\% confidence intervals.}
\end{figure}

The on-policy version of each algorithm performed better than the off-policy version in terms of overall performance.
Sarsa and $Q(\sigma = 0.5)$ performed better in terms of final performance when not using off-policy correction.
In terms of initial performance, the off-policy versions of each algorithm performed better than their on-policy counterparts.
Figure \ref{fig:off_vs_on} shows the average performance over intervals of 50 episodes with corresponding $95\%$ confidence intervals computed using a t-distribution. 
The differences in initial and final performance between the off-policy and on-policy versions of each of these algorithms were statistically significant as evidenced by the non-overlapping confidence intervals.
Table \ref{tbl:off_vs_on} shows the overall performance of each algorithm with their corresponding sample standard deviation.
The third column of the table shows the p-value of a Welch's test for the difference between the performance of the on-policy and the off-policy versions of each of the algorithms.

\begin{table}[h]
  \caption{Overall performance of each algorithm with off-policy correction (off-policy) and without (on-policy). The results are averaged over 100 independent runs. The number in parenthesis corresponds to the sample standard deviation. The third column corresponds to the p-value of the Welch's test for the difference in performance between the off-policy and the on-policy algorithms.}
  \label{tbl:off_vs_on}
  \centering
  \begin{tabular}{lccccc}
    \toprule
    & \multicolumn{4}{c}{Overall Performance} & \\
    \cmidrule(r){2-5}
    Algorithm               & \multicolumn{2}{c}{Off-Policy} & \multicolumn{2}{c}{On-Policy} & Welch's Test \\
                            & Avg & SD & Avg & SD & P-Value \\
    \midrule
    Sarsa                   & -829.33 & 100.52  & -308.92 & 61.42   & $< 1 \times 10^{-5}$ \\
    $Q(\sigma = 0.5)$       & -580.18 & 44.31   & -305.45 & 43.81   & $< 1 \times 10^{-5}$ \\
    Decaying $\sigma$       & -485.13 & 33.66   & -294.87 & 48.19   & $< 1 \times 10^{-5}$ \\
    \bottomrule
  \end{tabular}
\end{table}

The results of this experiment partially support our initial hypothesis.
There was no adverse effect on the performance of each algorithm in terms of final and overall performance when not using off-policy correction.
In fact, it seems that, if done naively, off-policy correction can have an adverse effect on learning.
However, the initial performance was significantly improved when using off-policy correction.
This effect suggests that an algorithm that used off-policy correction only during early training and no correction afterwards could achieve better performance.
Henceforth, we will study Sarsa and $Q(\sigma)$ without off-policy correction.

\subsection{Comparison of n-Step Algorithms}

The parameter $n$ has been shown to trade-off between the bias and the variance of the estimate of the return \parencite{Kearns:2000:BEB:648299.755183}.
This effect has been observed in the linear function approximation case, but remains to be systematically studied in the deep reinforcement learning setup.

In this experiment we study the effect of the back up length $n$ on the performance of Retrace, $Q$-learning, Tree Backup, Sarsa, $Q(\sigma = 0.5)$, and decaying $\sigma$.
For Retrace, we truncated the importance sampling ratio at $1$.
For decaying $\sigma$, we used a linear decay of $0.002$ as in the previous experiment.
Based on results obtained in the linear function approximation case with $n$-step Sarsa \parencite{sutton2018}, we hypothesized that using a backup length greater than $1$ would result in better performance, but using too big of a value would have an adverse effect.
In order to test this hypothesis, we implemented a version of each of these algorithms for each value of $n$ in $\{1, 3, 5, 10, 20\}$.
The measure of performance was the same as in the previous experiment.

\begin{figure}[t]
  \centering
  \label{fig:nstep}
  \includegraphics[width=0.85\textwidth]{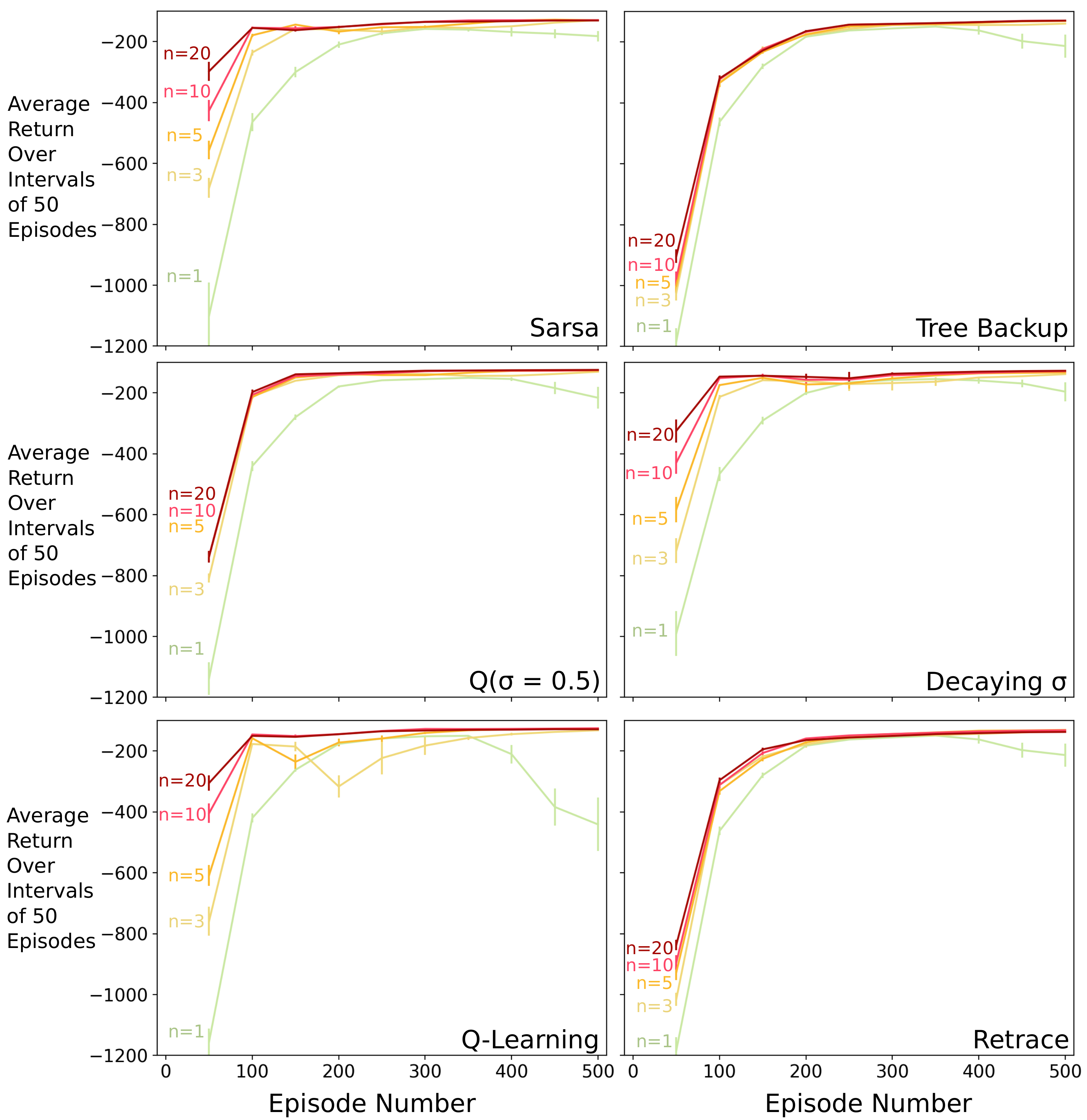}
  \caption{Average return over intervals of 50 episodes. The results are averaged over 100 independent runs. The error bars correspond to 95\% confidence intervals.}
\end{figure}

The $20$-step version of each algorithm performed better than with smaller values of $n$ in terms of initial and overall performance.
In terms of final performance, values of $n \geq 3$ performed similarly. 
The gain in overall performance was mainly due to improved initial performance.
This effect was of larger magnitude for $n$-step $Q$-learning, Sarsa, and decaying $\sigma$.
Figure \ref{fig:nstep} shows the average return over intervals of 50 episodes with corresponding $95\%$ confidence intervals.
More detailed summaries of the initial, final, and overall performance can be found in appendix A.

The results of this experiment partially support our initial hypothesis.
Increasing the value of $n$ improved the performance of the algorithm.
However, we were not able to find a value of $n$ that was too high and that, consequently, had an adverse effect in performance. 
Our results are reminiscent of previous work where $n$ was reported to be a sensitive hyper-parameter \parencite{rainbow2017}.
As evidenced by figure \ref{fig:nstep}, the value of $n$ can have drastic effects in performance.

\subsection{Target Network Update Frequency}

In preliminary experiments we noticed that algorithms that relied more heavily on bootstrapping, such as Tree Backup, were more sensitive to the target network update frequency.
We theorized that this effect was due to the mismatch in learning between the target network and the update network --- the network that is updated at every time step.
If this were true, then we would expect that $n$-step algorithms such as Tree Backup, Retrace, and $Q(\sigma)$ would be more sensitive to changes in the target network update frequency, whereas Sarsa and $Q$-learning would be more robust.
Note that this is the only experiment where we use a target network update frequency different from $1,000$.

To test this hypothesis we implemented the $20$-step versions of Retrace, Tree Backup, $Q(\sigma=0.5)$, Sarsa, $Q$-learning, and decaying $\sigma$ with target network update frequencies of 500, 1000, and 2000.
We used the average return over the whole training period as measure of performance.

\begin{table}[t]
  \caption{Overall performance of 20-step algorithms with target network update frequencies of 500, 1000, and 2000 time steps. 
  For each frequency we report the average (Avg) over 100 independent runs with its corresponding standard deviation (SD).
  The last two columns correspond to the difference (Diff) in performance between the algorithm with an update frequency of 500 and 2000; we report the p-value of the Welch's test to validate the results.
  }
  \label{tbl:tnetwork}
  \centering
  \begin{tabular}{lcccccccc}
    \toprule
    & \multicolumn{6}{c}{Update Frequency} &    \multicolumn{2}{c}{Performance Diff}              \\
    \cmidrule(r){2-7}
    & \multicolumn{2}{c}{500}   & \multicolumn{2}{c}{1000}  & \multicolumn{2}{c}{2000}  & \multicolumn{2}{c}{Freq 500 - Freq 2000} \\
    Algorithm  & Avg & SD & Avg & SD & Avg & SD & Diff & P-value \\
    \midrule
    Sarsa                   & -158.59 & 19.02   & -157.05 & 21.79   & -166.82 & 28.98 &  8.24   & $0.02$        \\
    $Q(\sigma = 0.5)$       & -164.32 & 3.72    & -196.79 & 7.71    & -257.83 & 13.88 &  93.51  & $< 10^{-5}$   \\
    Tree Backup             & -193.73 & 4.09    & -243.15 & 7.56    & -338.62 & 15.52 &  144.89 & $< 10^{-5}$   \\
    Decaying $\sigma$       & -148.46 & 19.63   & -156.97 & 40.48   & -178.46 & 63.38 &  30.0   & $< 10^{-5}$   \\
    Retrace                 & -193.11 & 4.16    & -235.83 & 7.48    & -310.73 & 14.22 &  117.62 & $< 10^{-5}$   \\
    $Q$-learning            & -151.77 & 18.45   & -153.80 & 18.03   & -160.55 & 32.3  &  8.78   & $0.03$        \\
    \bottomrule
  \end{tabular}
\end{table}

Increasing the target network update frequency had a more severe effect on the performances of Tree Backup, Retrace and $Q(\sigma = 0.5)$ than on the performances of Sarsa and $Q$-learning.
Table \ref{tbl:tnetwork} shows the results of this experiment averaged over 100 independent runs.

These results support our initial hypothesis.
Retrace, Tree Backup, and $Q(\sigma = 0.5)$ seem to be more sensitive to the value of the target network update frequency.
A possible explanation for this effect is that Retrace, Tree Backup, and $Q(\sigma = 0.5$ retrieve more estimates of the action value function in order to compute their target; in other words, they rely more heavily on the target network.

\section{Conclusion}

In this paper we have presented a systematic study of the performance of several multi-step algorithms combined with an architecture analogous to DQN \parencite{mnih2015humanlevel}.
We studied popular algorithms used in deep reinforcement learning --- Retrace, $Q$-learning, and Tree Backup --- as well as algorithms that had not been previously studied in the deep reinforcement learning setting --- Sarsa and $Q(\sigma)$.
Our statistical analysis allowed us to draw clear conclusions about the effect of the off-policy correction, the backup length parameter $n$, and the target network update frequency on the performance of each of these algorithms.

We presented three main results in this paper.
First, we found that it is possible to ignore off-policy correction without seeing an adverse effect in the overall performance of Sarsa and $Q(\sigma)$.
This finding is problem specific, but it suggests that off-policy correction is not always necessary for learning from samples from the experience replay buffer.
Nonetheless, off-policy correction seem to benefit performance during early training.
It seems possible to devise an algorithm that exploited the early performance of the off-policy correction without being affected by its adverse effect in late performance.
Second, we found that the parameter $n$ can result in significant improvement in the performance of all of these algorithms.
In this sense, our findings are reminiscent of previous results where $n$ was reported to be a sensitive hyper-parameter \parencite{rainbow2017}.
Finally, we found that some algorithms are more robust to the update frequency of the target network.
Our results seem to indicate that algorithms that retrieve more information from the target network are also more sensitive to changes in this hyper-parameter.
One interesting possibility is that these types of algorithms could greatly benefit from not using a target network at all, as long as it does not affect the stability of the network.


\printbibliography

\newpage

\section*{Appendix A: Initial, Final, and Overall Performance of n-Step Algorithms}

To supplement the results of the second experiment, we present the summaries of the initial, final, and overall performance of all the $n$-step methods.
The results are averaged over 100 independent runs and rounded to two decimal numbers.
Moreover, we report the $95\%$ confidence interval for each measure of performance to validate the results.

\begin{table}[h]
  \caption{Initial performance of $n$-step algorithms measured as the average return over the first 50 episodes. 
  The results are averaged over 100 independent run.
  We also provide the sample standard deviation, and the lower and upper $95\%$ confidence interval bounds to validate the results.
  }
  \label{tbl:initial_nstep}
  \centering
  \begin{tabular}{lccccc}
    \toprule
    & \multicolumn{5}{c}{Average Return Over the First 50 Episodes of Training}                   \\
    \cmidrule(r){2-6}
    Algorithm               & $n$ & Sample Average & Sample Standard Dev. & Lower C.I. & Upper C.I. \\
    \midrule
    \multirow{5}{*}{Sarsa}      & 1     & -1100.32  & 552.36    & -1209.92  & -990.72   \\
                                & 3     & -680.56   & 166.06    & -713.51   & -647.61   \\
                                & 5     & -556.17	& 152.59	& -586.45	& -525.9    \\
                                & 10    & -426.3	& 174.95	& -461.01	& -391.58   \\
                                & 20    & -297.2	& 157.85	& -328.52	& -265.88   \\
    \midrule
    \multirow{5}{*}{$Q(\sigma = 0.5)$}  & 1     & -1137.92	& 270.93	& -1191.68	& -1084.16  \\
                                        & 3     & -808.23	& 74.24	    & -822.97	& -793.5    \\
                                        & 5     & -737.34	& 73.57	    & -751.94	& -722.74   \\
                                        & 10    & -736.75	& 90.47	    & -754.71	& -718.8    \\
                                        & 20    & -737.67	& 97.03	    & -756.93	& -718.42   \\
    \midrule
    \multirow{5}{*}{Tree Backup}    & 1     & -1183.69	& 217.13	& -1226.77	& -1140.61  \\ 
                                    & 3     & -1023.8	& 127.66	& -1049.13	& -998.47   \\
                                    & 5     & -1000.27	& 129.58	& -1025.99	& -974.56   \\
                                    & 10    & -977.31	& 120.06	& -1001.13	& -953.49   \\
                                    & 20    & -903.61	& 111.75	& -925.79	& -881.44	\\
    \midrule
    \multirow{5}{*}{Decaying $\sigma$}  & 1  & -990.13	& 369.24	& -1063.4	& -916.86   \\
                                        & 3  & -718.19	& 207.22	& -759.31	& -677.08   \\
                                        & 5  & -583.33	& 211.94	& -625.38	& -541.28   \\
                                        & 10 & -428.97	& 186.41	& -465.96	& -391.98   \\
                                        & 20 & -324.7	& 190.43	& -362.48	& -286.91   \\
    \midrule
    \multirow{5}{*}{Retrace}    & 1  & -1183.69	& 217.13	& -1226.77	& -1140.61  \\
                                & 3  & -1014.93	& 108.81	& -1036.52	& -993.34   \\
                                & 5  & -926.82	& 127.74	& -952.17	& -901.48   \\
                                & 10 & -894.32	& 120.02	& -918.14	& -870.51   \\
                                & 20 & -836.39	& 85.93	    & -853.44	& -819.34   \\
    \midrule
    \multirow{5}{*}{$Q$-learning}   & 1  & -1155.34	& 216.73	& -1198.34	& -1112.34  \\
                                    & 3  & -758.8	& 241.47	& -806.71	& -710.89   \\
                                    & 5  & -608.56	& 174.37	& -643.16	& -573.96   \\
                                    & 10 & -404.38	& 162.39	& -436.6	& -372.16   \\
                                    & 20 & -304.97	& 130.06	& -330.77	& -279.16   \\
    \bottomrule
  \end{tabular}
\end{table}

Table \ref{tbl:initial_nstep} shows the summaries of the initial performance of each algorithm.
Just like reported in the results of the second experiment, across all the algorithms, increasing the value of $n$ consistently improved the initial performance.
This effect was of larger magnitude for Sarsa, $Q$-learning, and decaying $\sigma$.

\begin{table}[t]
  \caption{Final performance of $n$-step algorithms measured as the average return over the last 50 episodes. 
  The results are averaged over 100 independent run.
  We also provide the sample standard deviation, and the lower and upper $95\%$ confidence interval bounds to validate the results.
  }
  \label{tbl:final_nstep}
  \centering
  \begin{tabular}{lccccc}
    \toprule
    & \multicolumn{5}{c}{Average Return Over the Last 50 Episodes of Training}                   \\
    \cmidrule(r){2-6}
    Algorithm               & $n$ & Sample Average & Sample Standard Dev. & Lower C.I. & Upper C.I. \\
    \midrule
    \multirow{5}{*}{Sarsa}      & 1     & -181.95	& 85.95	& -199	    & -164.9    \\
                                & 3     & -129.47	& 9.61	& -131.38	& -127.56   \\
                                & 5     & -129.58	& 5.18	& -130.61	& -128.56   \\
                                & 10    & -129.11	& 5.8	& -130.26	& -127.96   \\
                                & 20    & -130.42	& 8.06	& -132.02	& -128.82   \\
    \midrule
    \multirow{5}{*}{$Q(\sigma = 0.5)$}  & 1     & -215.94	& 180.12	& -251.68	& -180.2    \\ 
                                        & 3     & -130.54	& 6.44	    & -131.81	& -129.26   \\ 
                                        & 5     & -126.85	& 4.63	    & -127.77	& -125.93   \\
                                        & 10    & -125.06	& 4.13	    & -125.88	& -124.24   \\
                                        & 20    & -124.34	& 3.34	    & -125  	& -123.68   \\
    \midrule
    \multirow{5}{*}{Tree Backup}    & 1     & -213.14	& 192.78	& -251.39	& -174.89   \\
                                    & 3     & -139.47	& 7.33	    & -140.92	& -138.01   \\
                                    & 5     & -130.99	& 3.99	    & -131.78	& -130.2    \\
                                    & 10    & -129.55	& 3.81	    & -130.3	& -128.79   \\
                                    & 20    & -129.47	& 4.16	    & -130.3	& -128.65   \\
    \midrule
    \multirow{5}{*}{Decaying $\sigma$}  & 1  & -195.84	& 157.31	& -227.05	& -164.63   \\
                                        & 3  & -138.38	& 10.04	    & -140.38	& -136.39   \\
                                        & 5  & -132.73	& 7.41	    & -134.2	& -131.26   \\
                                        & 10 & -128.67	& 4.89	    & -129.64	& -127.7    \\
                                        & 20 & -127.25	& 4.55	    & -128.15	& -126.34   \\
    \midrule
    \multirow{5}{*}{Retrace}    & 1  & -213.14	& 192.78	& -251.39	& -174.89   \\
                                & 3  & -134.29	& 6.41	    & -135.56	& -133.02   \\
                                & 5  & -130.7	& 4.98	    & -131.68	& -129.71   \\
                                & 10 & -131.83	& 4.4	    & -132.7	& -130.96   \\
                                & 20 & -137.28	& 5.74	    & -138.42	& -136.14   \\
    \midrule
    \multirow{5}{*}{$Q$-learning}   & 1  & -441	    & 441.97	& -528.7	& -353.31     \\
                                    & 3  & -131.98	& 9.39	    & -133.84	& -130.11     \\
                                    & 5  & -126.22	& 4.83	    & -127.18	& -125.26     \\
                                    & 10 & -125.24	& 4.32	    & -126.1	& -124.38     \\
                                    & 20 & -128.48	& 6.51	    & -129.77	& -127.19     \\
    \bottomrule
  \end{tabular}
\end{table}

Table \ref{tbl:final_nstep} shows the final performance of each $n$-step algorithm in terms of average return over the last 50 episodes of training.
For $n \geq 3$ most of the final performance was very similar for most of the algorithms. 
Retrace was the only algorithm that whose final performance peaked at an intermediate value of $n$.
Comparing all the algorithms in the table, $20$-step $Q(\sigma = 0.5)$ performed the best in terms of final performance.

\begin{table}[t]
  \caption{Overall performance of $n$-step algorithms as measure by the average return over the whole training period (500 episodes). 
  The results are averaged over 100 independent run.
  We also provide the sample standard deviation, and the lower and upper $95\%$ confidence interval bounds to validate the results.
  }
  \label{tbl:overall_nstep}
  \centering
  \begin{tabular}{lccccc}
    \toprule
    & \multicolumn{5}{c}{Average Return Over 500 Episodes of Training}                   \\
    \cmidrule(r){2-6}
    Algorithm               & $n$ & Sample Average & Sample Standard Dev. & Lower C.I. & Upper C.I. \\
    \midrule
    \multirow{5}{*}{Sarsa}      & 1     & -308.92	& 61.42	& -321.11	& -296.74   \\
                                & 3     & -308.92	& 61.42	& -321.11	& -296.74   \\
                                & 5     & -187.87	& 21.99	& -192.24	& -183.51   \\
                                & 10    & -168.43	& 25.06	& -173.4	& -163.45   \\
                                & 20    & -157.05	& 21.79	& -161.37	& -152.73   \\
    \midrule
    \multirow{5}{*}{$Q(\sigma = 0.5)$}  & 1     & -305.45	& 43.81	& -314.14	& -296.75   \\
                                        & 3     & -215.3	& 8.4	& -216.97	& -213.63   \\
                                        & 5     & -203.63	& 6.22	& -204.86	& -202.39   \\
                                        & 10    & -199.79	& 6.69	& -201.12	& -198.46   \\
                                        & 20    & -196.79	& 7.71	& -198.32	& -195.26   \\
    \midrule
    \multirow{5}{*}{Tree Backup}    & 1     & -314.76	& 39.24	& -322.54	& -306.97   \\
                                    & 3     & -262.66	& 9.28	    & -264.51	& -260.82   \\
                                    & 5     & -257.56	& 8	        & -259.15	& -255.97   \\
                                    & 10    & -250.57	& 8.4	    & -252.24	& -248.91   \\
                                    & 20    & -243.15	& 7.56	    & -244.65	& -241.65   \\
    \midrule
    \multirow{5}{*}{Decaying $\sigma$}  & 1  & -294.87	& 48.19	& -304.43	& -285.3    \\
                                        & 3  & -218.62	& 52.24	& -228.99	& -208.25   \\
                                        & 5  & -194.71	& 47.47	& -204.13	& -185.29   \\
                                        & 10 & -171.05	& 34.99	& -177.99	& -164.11   \\
                                        & 20 & -156.97	& 40.48	& -165	    & -148.94   \\
    \midrule
    \multirow{5}{*}{Retrace}    & 1  & -314.76	& 39.24	& -322.54	& -306.97   \\
                                & 3  & -259.4	& 11.47	& -261.68	& -257.12   \\
                                & 5  & -248.48	& 8.72	& -250.21	& -246.74   \\
                                & 10 & -240.39	& 8	    & -241.98	& -238.81   \\
                                & 20 & -235.83	& 7.48	& -237.31	& -234.34   \\
    \midrule
    \multirow{5}{*}{$Q$-learning}   & 1  & -350.88	& 78.82	& -366.52	& -335.24   \\
                                    & 3  & -241.36	& 77.51	& -256.74	& -225.98   \\
                                    & 5  & -198.77	& 37.22	& -206.16	& -191.39   \\
                                    & 10 & -161.34	& 21	& -165.5	& -157.17   \\
                                    & 20 & -153.8	& 18.03	& -157.38	& -150.23   \\
    \bottomrule
  \end{tabular}
\end{table}

Table \ref{tbl:overall_nstep} shows the overall performance measured as the average return over the whole training period (500 episodes).
Across all the algorithms, increasing the value of $n$ consistently improved the overall performance.
Even for the $n$-step Retrace algorithm, which best final performance is achieved with $n = 5$, the $20$-step version performed the best.
This seems to indicate that the improvement in overall performance is mostly due to improved initial performance.

The results found in these experiment are reminiscent of  what was reported in \citeauthor{rainbow2017} (\citeyear{rainbow2017}) where they found $n$ to be a sensitive hyper-parameter. 
If we take a look at the overall performance of $Q$-learning, for $n = 1,3,5$, the claim of $n$ being a ``sensitive'' hyper-parameter seems justifiable since the performance almost doubles when comparing $n = 1$ and $n = 5$. 
On the other hand, our results contrast the results from \citeauthor{rainbow2017} (\citeyear{rainbow2017}) since we found that our network performed better with a value of $n$ lot higher than the one used in the Rainbow architecture.

\end{document}